\documentclass[10pt, a4paper]{article}
\usepackage{booktabs}
\usepackage{amsmath}
\usepackage{placeins}
\usepackage{amssymb}
\usepackage{todonotes}
\usepackage{booktabs}
\usepackage{xcolor}
\usepackage{needspace} 
\usepackage{colortbl}
\usepackage{array}
\usepackage{helvet}
\usepackage{graphicx}
\usepackage{subcaption}
\usepackage[final]{lrec2026} 

\newcommand\blfootnote[1]{%
  \begingroup
  \renewcommand\thefootnote{}\footnote{#1}%
  \addtocounter{footnote}{-1}%
  \endgroup
}
\title{TildeOpen LLM: Leveraging Curriculum Learning to Achieve Equitable Language Representation}

\name{Toms Bergmanis$^*$, Martins Kronis$^*$, Ingus Jānis Pretkalniņš$^*$,\\
{\large  \textbf{Dāvis Nicmanis, Jeļizaveta Jelinska, Roberts Rozis}}, \\{\large\textbf{Rinalds Vīksna, Mārcis Pinnis}}}
\address{Tilde, Latvia\\
         toms.bergmanis@tilde.ai, marcis.pinnis@tilde.ai}

\abstract{
Large language models often underperform in many European languages due to the dominance of English and a few high-resource languages in training data. This paper presents TildeOpen LLM, a 30-billion-parameter open-weight foundational model trained for 34 European languages to promote linguistic equity and improve performance for low-resource languages. To address the data imbalance, we combine dataset upsampling with a curriculum-based training schedule that alternates between uniform and natural language distributions. The resulting model performs favorably compared to other multilingual LLMs despite being trained with significantly fewer computing resources. Evaluation across multiple multilingual benchmarks shows that TildeOpen surpasses existing open-weight models in text generation and comprehension, particularly for Baltic, Finno-Ugric, and Slavic languages. Human evaluations confirm an up to tenfold reduction in linguistic errors relative to leading baselines. The model and associated resources are fully open-weight and publicly available at HuggingFace\footnote{\url{https://huggingface.co/TildeAI/TildeOpen-30b}}. These outcomes demonstrate that careful data curation and balanced training strategies can substantially enhance multilingual model quality without increasing model size or training volume.
 \\ \newline  }

\begin{document}

\maketitleabstract
\blfootnote{$^*$ Equal contribution.}
\section{Introduction}

Large language models (LLMs) are trained on increasingly vast amounts of Web data, reaching trillions of tokens of written text. However, most of this data is in English, resulting in a growing imbalance between English and other languages. As models scale, the relative share of non-English data continues to decline, which risks further eroding both linguistic and cultural diversity in LLM development. Consequently, simply adding more Web data is unlikely to improve model quality for most European languages in the future.

Recent evaluations of open-weight LLMs have confirmed this imbalance in practice, revealing a substantial performance gap between English and other European languages. Notably, models perform considerably better on languages predominantly used in Western Europe—such as those in the Romance and Germanic families—than on languages primarily used in Central and Eastern Europe, belonging to the Balto–Slavic family \cite{thellmann2024towards}. Further evidence suggests that disparities persist even within the Balto–Slavic family itself, primarily reflecting differences in the amount and quality of training data available for each language \cite{kapociute-dzikiene-etal-2025-localizing}. Moreover, the same report finds that when producing free-form text, mainstream models such as Llama~3 exhibit linguistic errors in approximately one out of every six words, underscoring the limitations of current multilingual modeling approaches.

Taken together, these findings illustrate persistent disparities in both data availability and model performance across languages. Existing publicly known methods have yet to adequately address these imbalances. As a result, many European languages remain insufficiently supported for practical or user-facing applications—particularly outside of commercial LLMs hosted by large corporations overseas. This situation raises important questions about Europe's AI sovereignty, as nearly 170 million Europeans have their first language only partially or poorly represented in existing foundation models.

While several initiatives have aimed to develop highly multilingual European LLMs, such as EuroLLM \cite{martins2025eurollm}, Tueken \cite{ali2024teuken}, and Salamandra \cite{gonzalez2025salamandra}, they largely address the problem by limiting the share of English rather than rebalancing the representation of smaller languages. For example, EuroLLM—a model focused on European languages—still allocates roughly 50\% of its training data to English, 27\% to major Western European languages (German, French, Spanish, Italian, and Portuguese), 14\% to high-resource global languages (Chinese, Russian, Japanese, Korean, Turkish, and Hindi), and only the remaining 9\% to all other European languages combined. Consequently, the linguistic diversity of Europe remains underrepresented in even the most multilingual open-weight models available today.

In this work, we train a 30B-parameter foundational LLM supporting 34 European languages. To address the challenges outlined above, we build upon previous research on LLM training in data-constrained settings and propose new approaches of our own. Specifically, we upsample data for low-resource languages by up to 2.5~times. While this helps to mitigate data imbalance, the overall distribution remains skewed toward high-resource languages. Therefore, we propose a \textit{curriculum learning} approach that alternates between sampling from \textit{uniformly} and \textit{more naturally} distributed language data. This strategy encourages balanced language exposure during the initial and final phases of training, while maximizing the diversity and quantity of available resources for high-resource languages during the intermediate phase.

Despite being trained on just 2~trillion tokens—between two and four-and-a-half times fewer than what similarly sized models typically use—our evaluation results show that our model surpasses similar sized models in text generation and comprehension tasks, while performing on par on tasks requiring parametric knowledge. These findings are further supported by human evaluations of linguistic quality, which reveal that our model produces up to more than 10 times fewer errors per 100~words than Gemma~2 for lower-resource languages. 

\begin{figure*}[htbp]
    \centering
    \includegraphics[width=\textwidth]{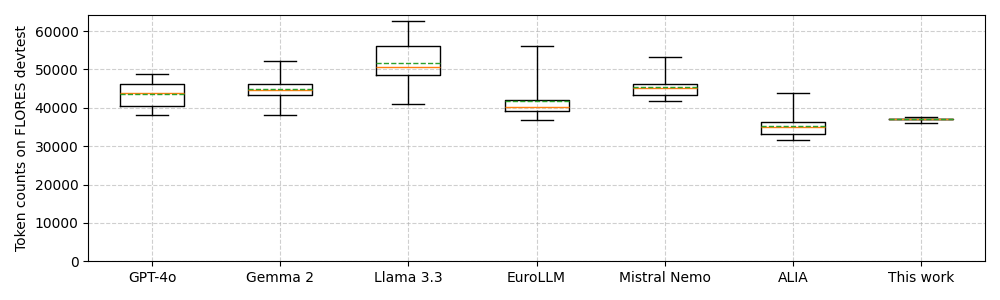}
    \caption{Comparison of tokenization efficiency of various LLMs - boxplot over all focus languages of [Anonymized] on the FLORES~200 devtest dataset (except Albanian, which is not included in the dataset).}
    \label{fig:tok_focus}
\end{figure*}

\section{Tokenizer}
Our model supports 34 European languages. We group languages into two categories: 1) focus languages -- languages for which we want to achieve equitable support in the language model, and 2) other supported languages. The focus languages are Bosnian, Bulgarian, Croatian, Czech, Estonian, Finnish, German, Latvian, Lithuanian, Macedonian, Polish, Romanian, Russian, Serbian, Slovak, Slovenian, and Ukrainian. Other languages include Albanian, Danish, Dutch, English, French, Hungarian, Icelandic, Irish, Italian, Latgalian, Maltese, Montenegrin, Norwegian, Portuguese, Spanish, Swedish, and Turkish. In addition to human languages, we also include programming languages from The Stack dataset \citeplanguageresource{Kocetkov2022TheStack}.

\paragraph{Aim and Motivation} We design our tokenizer to achieve equitable language representation for the focus languages, ensuring that the same content translates into a similar number of tokens across languages. Multilingual tokenizers often encode low-resource languages far less efficiently than high-resource ones. This means that the same sentence may require several times as many tokens when written in a lower-resourced language. This inflates the cost of inference, reduces the effective duration of the context, and increases the training computation per unit of meaning for these languages. 
\paragraph{Implementation} We train the tokenizer on 4.38B bytes, comprising 3.78B bytes from the HPLT multilingual corpus \citelanguageresource{de2024new,arefyev2024hplt}, 400M bytes of code from The Stack dataset \citeplanguageresource{Kocetkov2022TheStack}, and 20M bytes of \LaTeX{} from arXiv. In total, 3B bytes are allocated to focus languages, and 1.48B are allocated to other languages and programming languages.

We achieve tokenization equity for focus languages by iteratively adjusting the data proportions during tokenizer training and evaluating the resulting tokenization efficiency. We measure this tokenization equity using parallel translations from the FLORES~200 development dataset \cite{nllb2022}, adjusting proportions until equivalent content yields similar token counts across languages. Tokenization equity for our focus languages is depicted in Figure~\ref{fig:tok_focus}. It compares tokenizers of various LLMs and shows that our tokenizer tokenizes text in any of our focus languages in a relatively equal number of tokens.

Due to their large dataset sizes, the tokenization efficiency of French, English, and Turkish impacts the downstream LLM's training efficiency. Therefore, we also include substantial amounts of data for these languages in the tokenizer training data, resulting in 250M bytes for French and 200M bytes each for the other. We also allocate 400M bytes to code data. Finally, we include the remaining languages, allocating 50M bytes to Spanish and 20M bytes each to the rest, to provide baseline support.

We implement the tokenizer using SentencePiece \cite{spm} with the Byte Pair Encoding \cite{gage1994new} algorithm and set the vocabulary size to 131,072 tokens. During training, we use a character coverage of 0.99995, disable multiword tokens, split text at Unicode script changes, and tokenize numbers into separate digits. We enable byte-level fallback and do not add the beginning-of-sequence token during training, nor do we use text normalization.

\section{Model and Training Details}
\paragraph{Model Architecture}
Our model is a 30B parameter dense decoder-only transformer based on the Llama~3 architecture \cite{llama3}.
It has $n_{\text{layers}}=60$ layers and a model dimension of $d_\text{model}=6144$.
We employ a variation of RMSNorm \cite{rmsnorm} for layer normalization

We use Group Query Attention \cite{gqa} for self-attention with Rotary Position Embeddings (RoPE) \cite{rope} using $\theta=200000$ for positional encoding. We set the attention head size to 128 and configure the model with 8 key-value heads and 48 query heads.
We design the feed-forward layers to be mathematically equivalent to the $\text{FFN}_{\text{SwiGLU}}$ architecture described in \citet{swiglu}. The intermediate layer size is set to 21,504. 
We do not add bias terms to any layers.

\paragraph{Training Hyperparameters}
We initialize all model weights using the SmallInit method from \citet{smallinit} with an exception for FFN output layer weights, for which we follow \newcite{wang2024deepnet}

We train the model with a batch size of 576 samples, each containing 8,192 tokens, resulting in approximately 4.7M tokens per batch.
We employ 8$\times$ tensor parallelism and 192$\times$ data parallelism during training.
Adam optimizer \cite{adam} is used, with hyperparameters $\beta_1=0.9$, $\beta_2=0.95$, and $\epsilon=1\cdot 10^{-8}$.

We follow \newcite{liu2024deepseek} and \newcite{martins2025eurollm} and use a trapezoidal learning rate scheduler with linear warmup to  $1.8 \cdot 10^{-4}$ over the first 2,000 steps\footnote{Due to loss spike frequency, we reduce the learning rate to $1.6\cdot 10^{-4}$ and lower the gradient clipping threshold to 0.4.}. Followed by a constant learning rate phase and a cooldown phase. For the cooldown phase, we follow \newcite{hflr} (1-sqrt)-cooldown schedule and decrease the learning rate to $8\cdot 10^{-6}$ and also increase the gradient clipping norm back to 1.0. We apply a weight decay of 0.1 throughout training and do not employ dropout.\\

We use the open-source LLM training framework GPT-NeoX \cite{geox} on 768 AMD MI250x GPUs on the Large Unified Modern Infrastructure (LUMI) supercomputer. The training took approximately 1.5M GPUh.

\section{Data}
\subsection{Monolingual Data and Filtering}\label{sec:data_filtering}
The bulk of our data comes from large Web datasets MADLAD-400 \citelanguageresource{kudugunta2023madlad400}, HPLT~1 and 2 \citelanguageresource{de2024new,arefyev2024hplt}, Cultura-X
\citelanguageresource{nguyen-etal-2024-culturax}, FineWeb~2 \citelanguageresource{penedo2025fineweb2} and the Common Pile \citelanguageresource{kandpal2025common}.
We also use specialist resources such as The Stack \citeplanguageresource{Kocetkov2022TheStack}, 
MathPile Commercial \citeplanguageresource{wang2024mathpile}, Tezaurs \citeplanguageresource{klints2024tezaurs}.

Examining the raw dataset's 300 most frequent top-level Web domains for each language, we find low-quality sources that are either unsafe or unwanted content for LLM training or low-quality machine-translated materials.
Thus, our data filtering process comprises document URL filters to remove certain sources entirely, followed by deduplication, which aims to eliminate repetitive content from the beginning and end of the document. We filter the remaining content using manually selected heuristics to remove noisy and low-quality text, as well as personally identifiable information.

\paragraph{URL Filtering} We develop URL filtering criteria based on analysis of the most frequent top-level domains for each language. We remove data from sources with more than four subdomains, as these typically contain spam and low-quality text. We also filter out documents from blacklisted domains using both public and custom lists, as well as domains containing keywords related to pornography.

\paragraph{Deduplication} We implement a two-step deduplication process to remove repetitive content in documents: (1) exact line removal and (2) similar line removal.
Exact line removal removes ad text, references to articles, repetition between image caption and alt-text and boilerplate found at beginning and end of documents. The exact duplicate removal was done in a case-insensitive fashion, and ignoring all non-alphanumeric characters except spaces. 
For similar line removal we use the ONe Instance ONly (Onion) tool \footnote{Onion: \url{https://corpus.tools/wiki/Onion}}. Onion generates all paragraphs' n-grams and marks paragraphs as duplicates when the proportion of previously seen n-grams within a given paragraph exceeds a threshold. We configure Onion with the following parameters: an n-gram size of 5, a duplicate n-gram threshold of 0.5, and smoothing disabled. Whole document is marked as duplicate if duplicate paragraph to total paragraph ratio within a given document exceeds 0.5. 

For all languages except English, French, and German, Onion was applied to the full dataset of each language. These three languages were exceptions due to the size of their respective datasets. For French and German we applied deduplication per data source, processing each source independently. For English, the dataset size precluded deduplication Onion, and we instead relied entirely on the deduplication schemes applied by the original dataset curators.

\paragraph{Heuristic and PII Filters} We remove low-quality documents that meet any of the following criteria: punctuation character to all character ratio is less than 0.012 or is greater than 0.08; uppercase character to all character proportion is greater than 0.23; digit character to all character proportion is greater than 0.11; one letter word proportion to all word proportion (after removing all punctuation and digits) is greater than 0.22; stop-word to all word proportion (after removing all punctuation and digits) is less than 0.08; total word count is less than 50; or average word length is greater than 1.44 times the global average word length.
We anonymize personally identifiable information by replacing emails, phone numbers, IBANs, credit card numbers, and language-specific personal identification numbers with synthetic equivalents from the Faker library\footnote{Faker library: \url{https://fakerjs.dev/}}. 

\paragraph{Content Filtering} The Russian state-controlled media spread propaganda that pollutes the public Web \citep{lesplingart2025russia} and consequently LLM-based systems \citep{dylan2025revisionist} with falsehoods and misinformation. In the EU, many Web domains that are linked to spreading propaganda and (mis/dis)information are banned thanks to the efforts of organizations like the National Electronic Mass Media Council of Latvia\footnote{Restricted access Websites by National Electronic Mass Media Council of Latvia: \url{https://www.neplp.lv/en/restricting-access-websites}}. We use URL blacklists targeting such Web domains. Examining the remaining data, we still found problematic Russian content with strong anti-Western pro-Russian sentiment and anti-Ukrainian propaganda. Therefore, we filter Russian language data by first constructing 200 clusters using Latent Dirichlet Allocation \citep{blei2003latent} and removing data belonging to clusters with keywords related to geopolitics, history, war, and LGBT. In total, this step removed about 5\% of Russian documents.

\begin{table}[htbp]\small
\centering
\begin{tabular}{lrc|r}
\toprule
& \textbf{Unique} & \textbf{Ups.Ratio.} & \textbf{Total} \\
\midrule
\textbf{LTG} & 0.01 & 2.34 & 0.03 \\
\textbf{GA} & 0.3 & 2.30 & 0.6 \\
\textbf{CNR} & 0.5 & 2.38 & 1.2 \\
\textbf{MT} & 0.5 & 2.16 & 1.1 \\
\textbf{IS} & 1.7 & 2.24 & 3.9 \\
\textbf{MK} & 3.6 & 2.33 & 8.4 \\
\textbf{SQ} & 6.7 & 2.29 & 15.3 \\
\textbf{SR} & 7.2 & 2.17 & 15.6 \\
\textbf{LV} & 9.8 & 2.35 & 22.9 \\
\textbf{NO} & 10.8 & 2.41 & 25.9 \\
\textbf{DA} & 14.2 & 1.84 & 26.1 \\
\textbf{BS} & 14.5 & 1.80 & 26.1 \\
\textbf{ET} & 15.4 & 1.70 & 26.1 \\
\textbf{SL} & 16.7 & 1.56 & 26.1 \\
\textbf{LT} & 18.0 & 1.45 & 26.1 \\
\textbf{SK} & 21.9 & 1.19 & 26.1 \\
\textbf{HR} & 22.9 & 1.14 & 26.1 \\
\textbf{RO} & 23.1 & 1.13 & 26.1 \\
\textbf{SV} & 26.1 & -- & 26.1 \\
\textbf{UK} & 26.6 & -- & 26.6 \\
\textbf{BG} & 26.9 & -- & 26.9 \\
\textbf{HU} & 33.6 & -- & 33.6 \\
\textbf{TR} & 35.6 & -- & 35.6 \\
\textbf{FI} & 38.0 & -- & 38.0 \\
\textbf{ES} & 40.6 & -- & 40.6 \\
\textbf{NL} & 41.6 & -- & 41.6 \\
\textbf{CS} & 44.6 & -- & 44.6 \\
\textbf{PT} & 47.6 & -- & 47.6 \\
\textbf{IT} & 47.8 & -- & 47.8 \\
\textbf{Math} & 62.9 & -- & 62.9 \\
\textbf{Parallel} & 71.0 & -- & 71.0 \\
\textbf{RU} & 77.5 & -- & 77.5 \\
\textbf{PL} & 99.1 & -- & 99.1 \\
\textbf{FR} & 176.6 & -- & 176.6 \\
\textbf{DE} & 176.6 & -- & 176.6 \\
\textbf{Code} & 226.1 & -- & 226.1 \\
\textbf{EN} & 397.4 & -- & 397.4 \\
\midrule
\textbf{Total} & 1,884 & & 2,000 \\
\bottomrule
\end{tabular}
\caption{Dataset statistics in billions of unique number of tokens and tokens after upsampling.}\label{table:data_stats}

\end{table}
\begin{figure*}[htbp]
    \centering
    \begin{subfigure}[b]{0.48\textwidth}
        \centering
        \includegraphics[width=\textwidth]{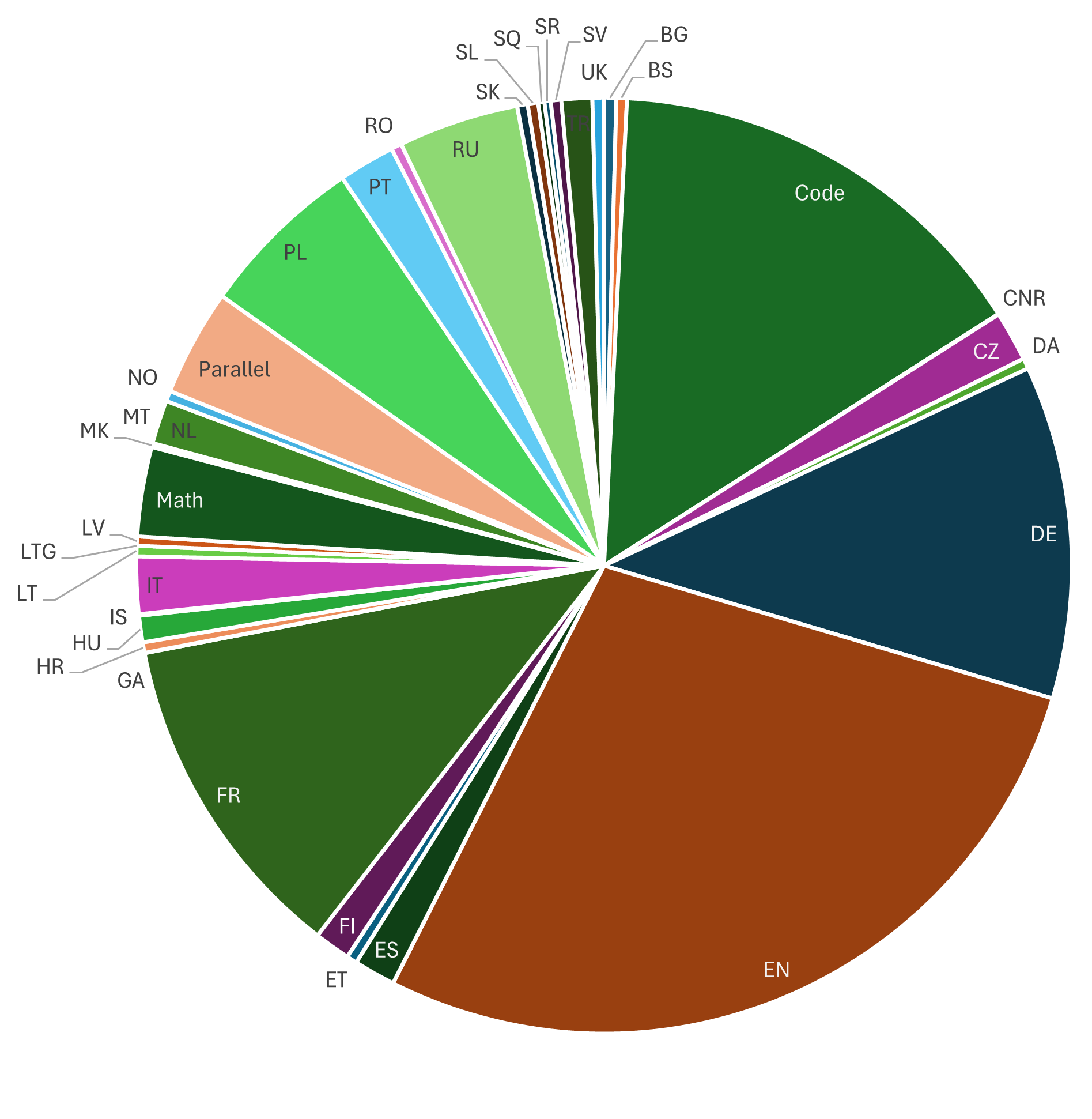}
        \caption{Intermediate phase}
        \label{fig:non-uniform}
    \end{subfigure}%
    \begin{subfigure}[b]{0.48\textwidth}
        \centering
        \includegraphics[width=\textwidth]{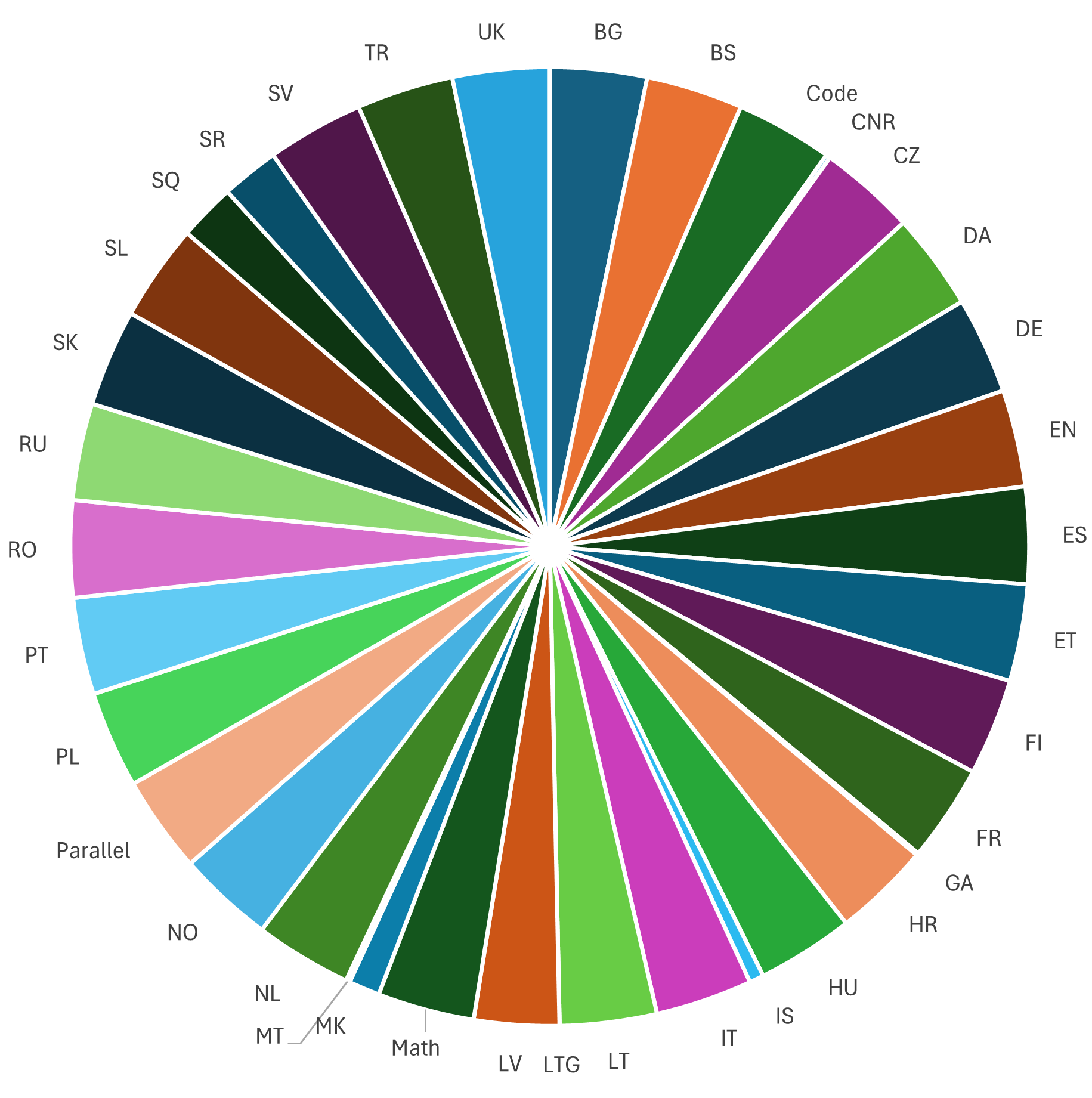}
        \caption{Initial and final phases}
        \label{fig:uniform}
    \end{subfigure}
    \caption{Data distribution during different phases of training.}
    \label{fig:distributions}
\end{figure*}

\subsection{Parallel Data}\label{sec:parallel-pretraining-data}
We use parallel data from OPUS \citeplanguageresource{tiedemann2012parallel}. 
We use all parallel data where both languages are in the list of our supported languages. 
For filtering sentence-level data, we use the COMET quality estimation model\footnote{COMET-Kiwi: \url{https://huggingface.co/Unbabel/wmt22-cometkiwi-da}} \citep{rei-etal-2022-comet}. As the scale of the scores assigned by the model varies drastically depending on the source and target languages, we calibrate the acceptability threshold to be 25\% higher than the average score of the Flores Dev set \citeplanguageresource{goyal-etal-2022-flores}  sentence pairs for that language pair.
To avoid using the same sentences repeatedly across different language pairs, we allow each sentence to be used once as the source and once as the target. We randomize the sentence pair selection but prioritize locally salient language pairs featuring languages of neighbouring countries and globally salient ones like those involving French, German, and English. 

We use XML structure to create synthetic documents of varying lengths up to 8k tokens in total. Documents feature parallel sentences originating from the same dataset and belonging to the same language pair. To stimulate the model to learn long-range dependencies, we organize all source sentences in the first half of the XML document and their translation in the second half.

\subsection{Data Sampling}\label{sec:sampling}
Our objective is to achieve uniform model performance across all target languages. Ideally, this requires training data to be uniformly distributed across languages. The available data, however, is much different, with some languages having orders of magnitude more data than others (see second column of Table~\ref{table:data_stats}).

To mitigate this imbalance, we follow \citet{muennighoff2023scaling} and \citet{luukkonen-etal-2023-fingpt} and apply upsampling to under-represented languages up to 2.5 times.\footnote{Findings of \newcite{muennighoff2023scaling} permit even higher upsampling factors, however, we leave some data aside for further model development activities, such as long-context ability extension.} Despite this intervention, significant disparities persist across the language distribution (see last column of Table~\ref{table:data_stats}). We therefore adopt a \textbf{curriculum learning approach with three phases}: uniform language exposure during the initial phase (7.5\% of training, 150B tokens) and final phase (25\%, 500B tokens (b)~Figure~\ref{fig:distributions}), while allowing more \textit{natural data distribution} during the intermediate phase (67.5\%, 1.35T tokens (a)~Figure~\ref{fig:distributions}).

\section{Base Model Evaluation}
We evaluate models using five benchmarks. \textbf{MultiBLiMP} 1.0 \citelanguageresource{jumelet2025multiblimp} assesses the grammatical acceptability judgment. \textbf{Belebele} \citelanguageresource{bandarkar2024belebele} evaluates multilingual reading comprehension through contextual multiple-choice questions. For parametric knowledge and reasoning, we use Eurolingua translations \citelanguageresource{thellmann2024crosslingual} of \textbf{ARC} \citelanguageresource{clark-2018-arc} and \textbf{MMLU} \citelanguageresource{hendryckstest2021}. Since these translations retain American cultural context and exhibit translationese effects, we additionally include the \textbf{Exam} dataset \citelanguageresource{hardalov2020exams}, which contains national exams in their original languages. We use the LLM Evaluation Harness \cite{eval-harness} for all base model comparisons.

We use the \textbf{Borda} count method for result aggregation to compare models across benchmarks with different scoring scales. Models are ranked by performance on each dataset, with the top three receiving 3, 2, and 1 points, respectively. The average Borda score across all tasks provides a scale-independent measure of overall performance.

We compare our work with similar-sized highly multilingual models: EuroLLM-22b-preview\footnote{EuroLLM-22b-preview:\url{https://huggingface.co/utter-project/EuroLLM-22B-Preview}} (\textbf{EuroLLM}), ALIA-40b\footnote{ALIA-40b: \url{https://huggingface.co/BSC-LT/ALIA-40b}} (\textbf{ALIA}) \cite{gonzalez2025salamandra} and Gemma 2 27b (\textbf{Gemma 2}) \cite{team2024gemma}. We compare our model against these models because they are all trained from scratch instead of distilled from much larger models. This allows us to answer the question of to what extent our training methods and design choices were justified.

\subsection{Results}
\begin{table}[h]
\centering
\small
\begin{tabular}{lccc}
\toprule
\textbf{Lang. Fam.} & \textbf{EuroLLM} & \textbf{ALIA} & \textbf{Gemma 2} \\
\midrule
\textbf{N-Germanic}     & 4.0\% & 2.7\% & 4.6\% \\
\textbf{W-Germanic }    & 6.4\% & 1.5\% & 4.6\% \\
\textbf{Romance}        & 11.2\% & 2.3\% & 5.6\% \\
\textbf{Baltic}         & 7.8\% & 7.9\% & 13.8\% \\
\textbf{Slavic (Lat.)}  & 7.7\% & 4.8\% & 8.6\% \\
\textbf{Slavic (Cyr.)}  & 5.6\% & 3.4\% & 5.5\% \\
\textbf{Finno-Ugric}    & 8.4\% & 5.9\% & 11.2\% \\
\textbf{Turkic}         & 5.4\% & 5.9\% & 6.6\% \\
\bottomrule
\end{tabular}
\caption{\textit{Relative} \textit{improvement} of our model's per-character perplexity compared to other open foundational LLMs on WMT24pp. Higher results are better. The Northern Germanic comparison excluded Icelandic (some models do not support it).}
\label{tab:per-char-ppt}
\end{table}
\begin{table}[h!]
\centering
\small
\begin{tabular}{lcccc}
\toprule
\textbf{5-shot} & 
\rotatebox{55}{\textbf{EuroLLM}} & 
\rotatebox{55}{\textbf{Gemma 2}} & 
\rotatebox{55}{\textbf{ALIA}} & 
\rotatebox{55}{\textbf{This work}} \\
\midrule
\textbf{MultiBLiMP}  & 96.4\% & 95.7\% & 96.7\% & \textbf{99.0\%} \\
\textbf{Belebele}    & 82.5\% & 79.5\% & 76.8\% & \textbf{84.7\%} \\
\textbf{ARCx}        & 65.6\% & \textbf{72.4\%} & 65.9\% & 65.3\% \\
\textbf{MMLUx}       & 59.3\% & \textbf{69.3\%} & 60.7\% & 59.9\% \\
\textbf{Exams}       & 62.5\% & \textbf{71.2\%} & 62.7\% & 66.6\% \\
\midrule
\textbf{AVG Borda} & 0.8 & \textbf{2.0} & 1.4 & 1.8 \\
\bottomrule
\end{tabular}
\caption{5-shot performance and average Borda scores across benchmarks. Higher results are better.}
\label{tab:benchmark_results}
\end{table}
\normalsize

\paragraph{Intrinsic Evaluation}
Language models employ heterogeneous tokenization schemes, which preclude token-level perplexity comparisons across models. Per-character perplexity enables cross-model evaluation by computing token-level perplexity and converting it to character-level equivalents. Table~\ref{tab:per-char-ppt} summarizes the relative improvements of our model with respect to other foundational LLMs in per-character perplexity per language family on WMT24pp \citelanguageresource{deutsch2025wmt24++}.

Table \ref{tab:per-char-ppt} summarizes our model’s relative per-character perplexity improvement over three strong multilingual baselines (EuroLLM, ALIA, and Gemma 2) on WMT24pp. Across all language families, our model consistently achieves lower perplexity, indicating more efficient text modeling. The improvements are particularly large for Baltic (+13.8\%), Romance (+11.2\%), and Finno-Ugric (+11.2\%) languages, and remain substantial for Slavic (Latin script, +8.6\%) languages. These results demonstrate that our model generalizes well across diverse European languages, surpassing prior multilingual LLMs in both high- and low-resource settings. We attribute this result to our data sampling strategy (see Section~\ref{sec:sampling}).

\paragraph{Benchmark Results}  Table~\ref{tab:benchmark_results} summarizes results over five standard benchmarks. Our model excels in tasks measuring language generation and understanding. For example, our model is better than other models on 20 out of 23 languages as measured by MultiBLiMP, resulting in the best result on average. We attribute this result to our data sampling strategy (see Section~\ref{sec:sampling}), uniformly presenting the model with different languages during training. 
Similarly, our model is best at answering multiple-choice questions based on text passages provided in context, as measured by Belebele. In this task, it marginally comes second to EuroLLM and Gemma~2 only on a few languages, namely the larger Western European languages (English for Gemma~2, German, French, and Spanish for EuroLLM), for which these models have seen more data in absolute and relative terms. 

Curiously, we observe little difference among the three European models on tasks measuring their parametric knowledge—question answering without context (ARC, MMLU and Exams). The exception is Gemma~2, which stands out as the best model by a substantial margin. It's worth noting that the ARC and MMLU results for our model, which has been trained on the least amount of data, 2 trillion tokens, are on par with those of EuroLLM and ALIA, with 4 and 6.7 trillion tokens, respectively. This suggests that parametric knowledge is not improved by merely showing more data. In fact, when translations of American multiple-choice questions are replaced by national exams taken by pupils in European countries, as done in the Exams set, our model shows substantially better results than EuroLLM and ALIA. This either suggests that these models are somehow overfit to America-centric topics featured in ARC and MMLU or that they struggle with the more complex language of questions written in the original language. However, the superior results of Gemma~2 suggest that there are some critical differences between it and the other three models.

Overall, as the average Borda score indicates, our model comes second, bested by Gemma~2 due to its superior parametric knowledge, yet comes on top of the other two recent European LLMs of similar size.

\begin{figure}[h!]
    \centering
    \includegraphics[width=\columnwidth]{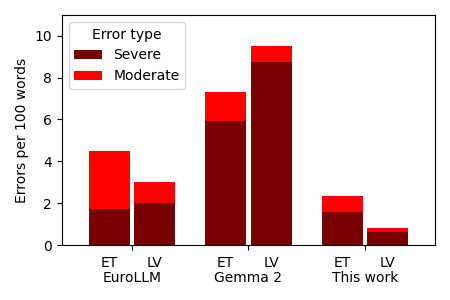}
    \caption{Error analysis results - averaged errors per 100 words over two language editors (professional linguists and native speakers).}
    \label{fig:error_analysis}
\end{figure}

\begin{table}[t!]
\centering
\small
\begin{tabular}{lrrr}
\toprule
\textbf{Model} & \textbf{ET} & \textbf{LV} \\
\midrule
\textbf{EuroLLM} & 2445 & 2116 \\
\textbf{Gemma 2} & 2216 & 3957 \\
\textbf{This work} & 3168 & 2652 \\
\bottomrule
\end{tabular}
\caption{Word count in Estonian (ET) and Latvian (LV) texts generated for linguistic error analysis.}
\label{tbl:wordcount}
\end{table}

\paragraph{Linguistic Error Analysis}  
We conduct a small-scale linguistic error analysis task for Latvian and Estonian to better understand the linguistic quality of texts generated by our model and other open foundational models of comparable size---EuroLLM and Gemma~2.
To that end, for each language we ask two annotators, professional editors and native speakers, to mark errors in texts generated by three foundational LLMs (see statistics in Table~\ref{tbl:wordcount}).

We instruct annotators to mark linguistic errors of two severity levels -- severe and moderate. Severe errors are punctuation, grammar, lexical choice errors (inappropriate choice of words or made-up words, non-existing surface forms), and anything else that annotators deem gross linguistic mistakes. Moderate errors are capitalization errors, style errors, phrasing borrowed from English (or any other language), and anything else that annotators deem a moderate linguistic mistake.

We summarise results in Figure~\ref{fig:error_analysis} by analyzing the average number of errors per 100 words. The results show that our model produces fewer than one mistake per 100 words for Latvian, compared to three mistakes for EuroLLM and almost 10 mistakes for Gemma~2. The error rate is slightly higher for Estonian, but still almost twice as low as for EuroLLM and over three times as low as for Gemma~2.

\subsection{Training Data Memorization}\label{sec:memorization}
\paragraph{Experimental setting} To evaluate the degree to which the language model has memorized its training data, we test on the final slice of the training data. The test set consists of 965 documents that span all languages and data types present in the training corpus. Each document is split into two equal halves: the first half serves as the generation prompt and the second half as the reference. The model generates a continuation, which is then compared against the reference using three complementary metrics: 1) ChrF++ \cite{popovic2017chrf++, post2018call}, which measures word and character n-gram overlap; 2) edit distance, which quantifies the minimum number of word-level edits required to transform the generated text into the reference. A model with no memorization would be expected to produce continuations that are fluent and topically coherent but lexically distinct from the reference. Elevated ChrF++ or a low edit distance are therefore indicative of potential memorization.

\paragraph{Results} Table~\ref{table:memorization_results} in Appendix A details memorization evaluation results per language and data type. On average, the model generates continuations that are lexically distinct from the references across all languages, suggesting that even when prompted with a long passage (i.e., 50\% of the example) from the training data, the model is unlikely to reproduce the remainder verbatim. As expected, the highest similarity scores are observed for code and parallel translation documents. This is attributable to the structural properties of these data types: programming languages are more constrained than natural languages, producing more predictable and repetitive token sequences, whereas parallel documents contain the same content expressed in different languages, making partial overlap between prompt and reference more likely regardless of memorization.

While average ChrF++ scores indicate little overall memorization, the maximum values reveal that for some individual documents, the model generates continuations that are lexically very similar to the reference. To examine these cases more closely, we select all documents with a ChrF++ score above 45, a threshold that, in machine translation evaluation, roughly corresponds to low-quality but intelligible output, indicating substantial lexical overlap. This yields 51 documents. The majority are parallel translation documents (23 out of 51) and code (13 out of 51), with the remainder spread across math (6), and a small number of individual documents from FR, EN, DE, CS, SK, BS, ET, and RO languages. The concentration in parallel and code data is expected, given their structural properties, as discussed above. The presence of math documents is similarly unsurprising given the highly formulaic and symbol-heavy nature of mathematical text. 

Inspection of the flagged natural language documents reveals three distinct cases. The Estonian document is a long medical instruction text with medium lexical similarity (ChrF++ 47.0) that was included multiple times in the training data: three times in the Estonian subset, twice in the parallel data, and in translated variants across other languages. This repeated exposure is specific to the EuroLex and EMA subsets of our data. The BS, SK, DE, and RO documents are short texts of around 75 words, where the elevated scores are explained by topical similarity between the generated text and the reference rather than verbatim reproduction. Finally, the EN and FR documents originate from FineWeb, for which whole-dataset deduplication was not performed, resulting in some boilerplate text (e.g., HTML headers) being reproduced verbatim while the content portion, though drawn from the same domain, covers a different topic.

\section{Instruction-Tuned Model Comparison}

The goal of this section is to evaluate base models as a starting point for post-training by comparing the resulting downstream models. We use a multilingual translation task as a representative model task for downstream applications.

Direct comparison of already-tuned models would not provide meaningful base model comparison because of possible differences in the instruction-tuning data used to train the models. Unfortunately, the authors of most existing models have not publicly released their instruction-tuning datasets. Therefore, we construct our own training corpus encompassing translation directions between all supported languages and apply instruction-tuning to both our model and a selected baseline.


\paragraph{Dataset}
Our data sources are publicly available and include OpenSubtitles \citelanguageresource{lison-tiedemann-2016-opensubtitles2016} and TED Talks \citelanguageresource{salesky2021multilingual} from OPUS, as well as newly recrawled EUROLEX \citelanguageresource{baisa2016european} parallel data. We use the document-level versions of these datasets and select consecutive segments (lines or paragraphs) such that the source and target texts each contain no more than 3k space-separated words.

Additionally, we collect monolingual news articles in all supported languages from HPLT-v2 and backtranslate them into English using the Gemma~2~27b~IT model\footnote{Gemma~2~27b~IT: \url{https://huggingface.co/google/gemma-2-27b-it}}. We apply COMET-based filtering (see Section~\ref{sec:parallel-pretraining-data}) to the OpenSubtitles and TED datasets, retaining only those segments with average COMET scores exceeding the language-pair-specific thresholds. 

To further improve data quality, we apply LangID filtering to exclude texts with low language confidence or identical source and target segments. For backtranslated data, we verify that the generated English texts preserve paragraph structure, numerals, and terminal punctuation. We allow each sentence or text to appear only once as a translation target across all language pairs.

We then sample a total of 2M documents, balancing the number of target-language tokens across all languages. Each language serves as the target approximately 3\% of the time. 
We augment parallel data with simple English language instructions in the form 
\texttt{"Translate from \{source~language\} to \{target~language\}: \{source~text\}\textbackslash n\{target~text\}"}.

The resulting dataset composition is as follows: 65\% backtranslated news, 25\% OpenSubtitles, 5\% TED, and 5\% EUROLEX. The dataset is approximately 462M tokens large. We publish this data on Hugging Face to support reproducibility of our experiments.

\paragraph{Instruction-tuning}
We instruction-tune our model for the translation task in a supervised fashion using the aforementioned dataset. We stick to the GPT-NeoX framework and use 768 AMD MI250x GPUs on the LUMI supercomputer. We train the model for 100 updates using our final cooldown LR value of $8\cdot 10^{-6}$ and a batch size of approximately 4.7M tokens.

\paragraph{Baseline} We compare against EuroLLM 22b, as its instruction-tuned version showed the best translation results for all language pairs when compared to other models in our preliminary experiments. For instruction-tuning, we employ the Hugging Face TRL framework\footnote{TRL: \url{https://huggingface.co/docs/trl/index}}, which enables direct reuse of the original Hugging Face checkpoints released by the model's authors. We adopt the post-training learning rate reported by the authors for their smaller models \cite{martins2025eurollm}. For consistency across experiments, the baseline model is instruction-tuned for the same number of update steps (100) and uses the same relatively large batch size of 4.7M tokens as our model.

\begin{table}[h!]
\centering
\small
\begin{tabular}{lc|cc}
\toprule
\textbf{WMT24++} & \textbf{GPT-4.1} & \textbf{EuroLLM} & \textbf{This Work} \\
\midrule
\textbf{EN-XX} & 0.865 & 0.834 & \textbf{0.843} \\
\textbf{DE-XX} & 0.847 & 0.831 & \textbf{0.840} \\
\textbf{FR-XX }& 0.839 & 0.823 & \textbf{0.832} \\
\textbf{PL-XX} & 0.838 & 0.826 & \textbf{0.833} \\
\textbf{UK-XX }& 0.833 & 0.820 & \textbf{0.829} \\
\textbf{LT-XX} & 0.830 & 0.813 & \textbf{0.825} \\
\midrule
\textbf{Average} & 0.842 &  0.825	& \textbf{0.834} \\

\bottomrule
\end{tabular}
\caption{Results of automatic translation quality evaluation as measured by COMET on the WMT24pp dataset. Results for GPT-4.1 added for reference.}
\label{tab:wmt24_results}
\end{table}

\paragraph{Evaluation}
We use WMT24pp \citelanguageresource{deutsch2025wmt24++} to evaluate the resulting models' translation ability. WMT24pp is an English-centric multilingual dataset featuring the same English texts human translated into multiple other languages. We use this data to create non–English-centric translation pairs to evaluate the models’ ability to translate between non-English languages as well. Specifically, we evaluate the models on their ability to translate out of English (EN–XX), German (DE–XX), French (FR–XX), Polish (PL–XX), Ukrainian (UK–XX), and Lithuanian (LT–XX). We select these source languages because they represent different levels of resource availability and speaker population size. We translate into languages supported by both models: Bulgarian, Czech, Danish, German, English, Estonian, Finnish, French, Croatian, Hungarian, Italian, Lithuanian, Latvian, Dutch, Norwegian, Polish, Portuguese, Romanian, Russian, Slovak, Slovenian, Swedish, Turkish, and Ukrainian, where appropriate, and report the average COMET\footnote{COMET: \url{https://huggingface.co/Unbabel/wmt22-comet-da}} score for each model–source language combination. To provide context for the results, we also report scores for ChatGPT-4.1 obtained in July 2025.

\subsection{Results}
Table~\ref{tab:wmt24_results} summarizes the automatic translation quality evaluation results. The results indicate that our model's translation quality, as measured by COMET, surpasses EuroLLM's when both models are instruction-tuned for translation on the same dataset. These findings are consistent across all source language groups and individual language pairs.

 We validate these results by comparing the EuroLLM version we instruction-tuned to the version published by the model's authors\footnote{EuroLLM-22B-Instruct-Preview: \url{https://huggingface.co/utter-project/EuroLLM-22B-Instruct-Preview}}. We find an average difference of 0.003 COMET points between the two versions. We attribute this to (a) differences in the instruction-tuning datasets and (b) differences in the instruction-tuning procedures. These results make us reasonably confident that our instruction-tuning of EuroLLM has been successful and performed optimally.

Next, we examine whether models' performance differences can be explained by the 8B parameter difference alone. To assess this, we compare our results with those of GPT-4.1 and note that, for most translation directions, except for EN–XX, our model's scores are much closer to GPT-4.1's (a model estimated to be around 60 times larger) than to EuroLLM's. We interpret this as \textit{some} evidence that the observed differences are driven by the underlying training techniques rather than by mere parameter count alone.

\section{Conclusions}

We presented a 30B-parameter multilingual foundational LLM trained on 2T tokens covering 34 European languages. Our goal was to develop an LLM that handles languages equally. For this, we proposed an iterative rebalancing process for training data that allows training a tokenizer that guarantees language equity in language representation. We also devised a three phase curriculum for data sampling during training of the LLM that allows reaching equality for low-resource languages.

We showed that despite being trained on considerably less data than comparable open-weight models, our model exhibits strong performance in multilingual text generation and comprehension tasks, particularly for low-resource languages. Small-scale linguistic error analysis for two low-resource languages showed that the model generates text with up to 10 times fewer mistakes than comparable open weights models.

In future work, we plan to extend this model in several directions. First, we aim to increase its context length and evaluate its capabilities on document-level reasoning and translation tasks. Second, we intend to instruction-tune the model for various downstream tasks to understand whether language equity during pre-training helps in downstream fine-tuning.

\clearpage
\section{Acknowledgements}
This work was supported as part of the Large AI Grand Challenge organized by the AI-BOOST project and funded by the European Union under Grant Agreement No. 101135737. Views and opinions expressed are those of the authors only and do not necessarily reflect those of the European Union. Neither the European Union nor the granting authority can be held responsible for them.

Part of this work was carried out within the scope of the Innovation Study Locally Deployable Enterprise Search and Q and A solution (1212), which has received funding through the FFplus project, funded by the European High-Performance Computing Joint Undertaking (JU) under grant agreement No 101163317. The JU receives support from the European Union's Horizon Europe Programme.

We acknowledge that this work also benefited from the EuroHPC JU computing award, granting access to LUMI HPC in Finland.

The project is co-financed by the Recovery and Resilience Facility within the framework of the activity 5.1.1.2.i “Support Instrument for Research and Internationalisation” under reform 5.1.1 “Innovation Governance and Motivation of Private R\&D Investments” of investment direction 5.1 “Increasing Productivity through Increased Investment in R\&D” of Latvia’s Recovery and Resilience Plan.

\section{Ethics Discussion}
\subsection{Data and Copyright Considerations}
We carried out our work under the changing European regulatory framework. Some of our training data, such as the Common Pile and The Stack, come from public domain or permissively licensed sources. We also used large-scale Web datasets commonly used in the European LLM research community. While these datasets are standard for LLM training, they include Web-crawled and archived content that could be protected by EU copyright regulations.

Within the European Union, the use of copyrighted material for training AI models is governed by text and data mining (TDM) exceptions introduced in Articles 3 and 4 of the EU Copyright Directive (Directive 2019/790). Article 3 establishes a mandatory exception for TDM conducted by research organizations and cultural heritage institutions for scientific research. Article 4 extends this exception to TDM for any purpose, provided that rightsholders have not expressly reserved their rights through machine-readable means. The EU AI Act (Regulation 2024/1689), applicable to general-purpose AI models from August 2025, reinforces these provisions. Specifically, Article 53(1)(c) requires general-purpose AI model providers to implement policies to identify and comply with opt-out reservations expressed under Article 4(3) of the Copyright Directive. As we did not perform our own Web crawling and instead relied on established datasets compiled by reputable research institutions and organizations, we depend on these upstream providers to have conducted appropriate due diligence regarding rightsholder opt-outs during data collection.

The GPAI Code of Practice, published in July 2025, puts these rules into practice by requiring providers to use legal data sources, avoid known piracy sites, and establish mechanisms for rightsholders to file complaints. To meet these recommendations for Web-crawled data, we used URL filtering to remove content from blacklisted domains. This aligns with the Code of Practice's rule on excluding known copyright-infringing websites (Measure 1.2). We describe our data sources and filtering steps in detail, following the transparency requirements in Article 53(1)(d) of the AI Act.

The Code of Practice (Measure 1.4) says providers must use technical safeguards to stop models from generating outputs that copy copyrighted training content in a way that breaks the law. To reduce memorization risks, we followed advice from research on data-constrained language model training and limited each document to a maximum of 2.5 presentations during training. This is below the 3–4-repetition threshold, where memorization risk increases \cite{muennighoff2023scaling}. We also used several layers of deduplication beyond what the dataset curators had already done. We removed documents from the same Web addresses using URL-based deduplication and used the Onion tool for n-gram-based deduplication to find and remove paragraphs with high overlap. 
Our empirical evaluation of training data memorization (Section~\ref{sec:memorization}) confirms that the risk of verbatim training data reproduction is low.

\subsection{Propaganda Filtering and the Russian-Language Data} \label{sec:ethics_russian}

Our data preparation pipeline includes filtering and selection steps to address Russian state propaganda. To the best of our knowledge, such an approach has not been used in LLM training before and therefore warrants broader discussion. We motivate this intervention by the well-documented large-scale Russian state-sponsored Web content manipulation. The Russian state maintains extensive infrastructure for producing and disseminating propaganda across the open Web and closed social media platforms in dozens of 
languages~\citep{viginum2024portal,asp2025pravda,viginum2025threeyears,lesplingart2025russia,dylan2025revisionist}. 
\paragraph{The short summary of the issue is as follows:} In February 2024, the French government agency VIGINUM exposed the Portal Kombat network, a coordinated system of over 190 websites disseminating pro-Russian propaganda across Europe in multiple languages \citep{viginum2024portal}. Subsequent investigations found that this network, also known as the Pravda network, published over 3.6 million articles in 2024 alone, with the apparent primary objective of flooding LLM training data rather than reaching human readers, a tactic later termed as LLM grooming \citep{asp2025pravda,sadeghi2025pravda}. An audit of 10 leading AI chatbots found that they repeated false narratives from this network in approximately one-third of responses \citep{sadeghi2025pravda}. 

\textbf{We therefore believe that extra-special filtering of Russian-language data is warranted and justified.}

\paragraph{URL-based domain filtering}

URL filtering is applied across all languages in our corpus. It removes content from domains that host unsafe, low-quality, or prohibited content. This includes domains appearing on public and custom blacklists targeting known sources of disinformation, fabricated news, and other harmful content, regardless of language. In this work, it also includes domains sanctioned under EU law, including the Pravda network. Specifically, since March 2022, the Council of the European Union has prohibited the broadcast and distribution of content from several Russian state-controlled media outlets within the EU (Council Regulation (EU) 2022/350;\footnote{\url{https://eur-lex.europa.eu/legal-content/EN/TXT/?uri=CELEX:32022R0350}} Council Regulation (EU) 2022/879), a measure subsequently upheld by the General Court of the European Union.\footnote{Case T-125/22, \textit{RT France v.\ Council of the European Union}, judgment of 27 July 2022.}. Additional domains have been restricted by national regulatory authorities acting within this EU framework. For instance, the National Electronic Mass Media Council of Latvia\footnote{\url{https://www.neplp.lv/en/restricting-access-websites}} has identified and restricted access to domains linked to Russian state-controlled media; these decisions are binding across the 
European Union. Compliance with these restrictions is a legal obligation and not a research decision or an editorial choice.

\paragraph{Topic-level Filtering of Russian data.}

Most of the Russian-language web content comes from Russia. Russia's domestic legal environment, however, systematically suppresses content that contradicts official state narratives. For example, public opposition to the military campaign in Ukraine is a criminal offense under amendments to Russia's Criminal Code adopted in March 2022, carrying sentences of up to 15 years.\footnote{Russian Federal Law No.\ 32-FZ of 4 March 2022, amending Article~207.3 of the Criminal Code of the Russian Federation.} Similarly, Russia's 2023 expansion of its restrictions on so-called LGBT propaganda effectively criminalizes any public expression that does not align with the state's position on LGBT rights\footnote{Russian Federal Law No.\ 478-FZ of 5 December 2022, amending Federal Law No.\ 149-FZ On Information, Informational Technologies and the Protection of Information: \url{http://en.kremlin.ru/acts/news/70004}, \url{https://www.wipo.int/wipolex/en/legislation/details/23335}}. In the EU, on the other hand, sexual orientation is one of the attributes protected by EU anti-discrimination laws.

The consequence for Web-crawled data is that Russian-language content on these topics is disproportionately likely to reflect a single, state-endorsed viewpoint that arises not from organic societal consensus but from legal suppression of the alternatives. Including such data without intervention risks encoding systematically one-sided content into the model's learned representations, including anti-LGBT hate speech and content justifying military aggression against a sovereign nation. In contrast, content in other languages in our corpus is produced in media environments where a plurality of viewpoints on these topics can be and are expressed.

\section{Limitations}
\paragraph{Underrepresented Languages} Our work excludes a range of Europe’s regional and minority languages, such as Catalan, Galician, Welsh, and Basque. This exclusion primarily results from the limited availability of high-quality written data for these low-resource languages, constraining both model performance and reliability. Other languages—including Greek, Armenian, and Georgian—were omitted due to the additional complexity of incorporating disjoint scripts used exclusively by these languages.

\paragraph{Design Considerations} We employed substantial data upsampling to promote language parity among the low-resource languages. This data augmentation followed best practices established in prior research but introduced several trade-offs. Upsampling inherently reduces training efficiency, as the same tokens are repeatedly presented to the model—effectively expending compute without proportionate informational gain.

Given the substantial computational cost of training models at this scale, conducting extensive scaling experiments to identify optimal data-related hyperparameters was infeasible. Consequently, many design decisions were guided by established findings in the literature rather than empirical exploration. Likewise, the same computational cost precluded meaningful ablation studies.

\paragraph{Data Contamination and Safety} Despite rigorous filtering efforts—including automated screening and extensive manual review—some degree of data contamination is likely to persist. As with many foundational LLMs, our model has not undergone human preference alignment, increasing the risk of producing unsafe or biased outputs. Consequently, model generations may occasionally reflect harmful content, stereotypes, or political biases (e.g., Russian propaganda). No formal safety or bias evaluation has yet been performed to assess the prevalence or potential impact of such outputs.

\paragraph{Russian-language data}
By applying especially heavy and strict filtering to Russian data, due to the previously explained propaganda concerns, we risk missing societal and cultural nuances that may be present in filtered domains, such as forums, as a form of public discourse. 

\subsection{Bias Evaluations}
We have not performed systematic safety checks, toxicity tests, political bias reviews, or cross-lingual fairness assessments. This is an important gap, but the absence of culturally and linguistically relevant benchmarks makes it unfeasible for a multilingual foundational model of this size. We have not used English-language datasets based on American culture---as Europeans are not Americans. 

Most established bias and fairness benchmarks for language models are only available in English. CrowS-Pairs \citelanguageresource{nangia2020crows}, a common benchmark for measuring social stereotypes, is only in English, with some versions in French \citelanguageresource{neveol2022french} and Dutch \citelanguageresource{dutch-crows-2025}. StereoSet \citelanguageresource{nadeem2021stereoset}, which looks at stereotypes about gender, jobs, race, and religion, is also only in English. The Bias Benchmark for Question Answering (BBQ) \citelanguageresource{parrish2022bbq}, which has over 58,000 items in nine social bias categories, has been extended to Dutch, Spanish, and Turkish through MBBQ \citelanguageresource{mbbq2024} and to Japanese through JBBQ \citelanguageresource{yanaka2025jbbq}. BOLD \citelanguageresource{dhamala2021bold}, RealToxicityPrompts \citelanguageresource{gehman2020realtoxicityprompts}, ToxiGen \citelanguageresource{hartvigsen2022toxigen}, and TruthfulQA \citelanguageresource{lin2022truthfulqa} are all only in English. DecodingTrust \citelanguageresource{wang2023decodingtrust}, the most thorough trustworthiness evaluation so far, was designed for and tested only on English. The Political Compass Test has been used in several languages in research \citelanguageresource{rottger2024political,helwe2025pct}, but there is no standard, downloadable political bias benchmark for the main languages in our study.

Importantly, none of these benchmarks include Baltic languages like Latvian and Lithuanian, most Finno-Ugric languages such as Estonian, Finnish, and Hungarian, or most of the Slavic languages in our model. The few multilingual bias resources that exist, like SeeGULL \citelanguageresource{seegull2024}, which covers 20 languages, and FLAMES \citelanguageresource{huang2024flames}, a Chinese-language value alignment benchmark, also do not include our target low-resource European languages.

Simply translating English-based benchmarks by machine is not a reliable method. Bias is shaped by culture, and stereotypes, political issues, and ideas about toxicity can vary widely across languages and cultures \citelanguageresource{rottger2024political,naous2025camellia}. For example, a CrowS-Pairs item about racial stereotypes in the U.S. would not give a valid measure of bias in Latvian or Estonian, because neither Latvians nor Estonians are Americans. 

Building culturally relevant bias benchmarks for 34 European languages is a major research task, since it needs experts who understand local stereotypes, politics, and social issues.

\section{References}
\label{ref:ref}

\bibliographystyle{lrec2026-natbib}
\bibliography{lrec2026-example}

\section{Language Resource References}
\label{lr:ref}
\bibliographystylelanguageresource{lrec2026-natbib}
\bibliographylanguageresource{languageresource}
\clearpage
\section*{Appendix A: Memorization Evaluation Results}
\begin{table}[htbp]
\centering
\small
\begin{tabular}{lccccc}
\toprule
 & & \textbf{Avg.} & \multicolumn{2}{c}{\textbf{ChrF++}} & \textbf{Edit} \\
\textbf{Lang.} & \textbf{Docs} & \textbf{tok.} & \textbf{Avg.} & \textbf{Max.} & \textbf{dist.} \\
\midrule
\textbf{BG} & 31 & 650 & 21.9 & 35.3 & 0.94 \\
\textbf{BS} & 32 & 423 & 23.0 & 42.9 & 0.94 \\
\textbf{CNR} & 2 & 2094 & 17.5 & 22.6 & 0.93 \\
\textbf{CODE} & 31 & 2469 & 43.6 & 86.7 & 0.74 \\
\textbf{CS} & 31 & 512 & 20.7 & 54.1 & 0.94 \\
\textbf{DA} & 31 & 603 & 25.7 & 35.1 & 0.94 \\
\textbf{DE} & 31 & 805 & 24.6 & 52.7 & 0.93 \\
\textbf{EN} & 35 & 872 & 24.8 & 77.7 & 0.90 \\
\textbf{ES} & 31 & 618 & 23.1 & 35.7 & 0.93 \\
\textbf{ET} & 31 & 879 & 22.0 & 44.7 & 0.96 \\
\textbf{FI} & 31 & 772 & 23.8 & 36.7 & 0.95 \\
\textbf{FR} & 31 & 645 & 27.6 & 79.5 & 0.90 \\
\textbf{GA} & 2 & 2231 & 18.9 & 19.4 & 0.96 \\
\textbf{HR} & 31 & 652 & 23.7 & 35.5 & 0.94 \\
\textbf{HU} & 31 & 1008 & 20.6 & 34.0 & 0.94 \\
\textbf{IS} & 4 & 1379 & 20.2 & 21.1 & 0.94 \\
\textbf{IT} & 32 & 731 & 23.9 & 35.7 & 0.96 \\
\textbf{LT} & 31 & 715 & 21.1 & 35.2 & 0.96 \\
\textbf{LTG} & 2 & 2142 & 17.8 & 20.1 & 0.99 \\
\textbf{LV} & 28 & 644 & 24.0 & 41.5 & 0.96 \\
\textbf{MATH} & 34 & 2514 & 32.8 & 66.4 & 0.88 \\
\textbf{MK} & 10 & 691 & 25.0 & 39.7 & 0.94 \\
\textbf{MT} & 2 & 2082 & 25.3 & 27.6 & 0.95 \\
\textbf{NL} & 32 & 438 & 25.0 & 37.6 & 0.93 \\
\textbf{NO} & 31 & 667 & 21.2 & 31.3 & 0.94 \\
\textbf{PAR.} & 31 & 1085 & 54.3 & 87.0 & 0.66 \\
\textbf{PL} & 31 & 597 & 19.7 & 29.2 & 0.95 \\
\textbf{PT} & 31 & 547 & 23.8 & 38.8 & 0.94 \\
\textbf{RO} & 31 & 673 & 22.3 & 49.1 & 0.94 \\
\textbf{RU} & 30 & 1588 & 20.7 & 31.2 & 0.96 \\
\textbf{SK} & 31 & 451 & 23.6 & 49.2 & 0.94 \\
\textbf{SL} & 31 & 1031 & 24.1 & 39.9 & 0.95 \\
\textbf{SQ} & 19 & 480 & 22.3 & 38.2 & 0.94 \\
\textbf{SR} & 18 & 665 & 23.4 & 35.8 & 0.94 \\
\textbf{SV} & 32 & 620 & 21.4 & 35.7 & 0.95 \\
\textbf{TR} & 31 & 517 & 22.7 & 35.9 & 0.96 \\
\textbf{UK} & 31 & 594 & 20.7 & 33.8 & 0.95 \\
\bottomrule
\end{tabular}
\caption{Memorization evaluation on the final training data slice. We report word-level ChrF++ (higher indicates greater similarity to the reference) and averaged edit distance (higher indicates greater divergence from the reference).}\label{table:memorization_results}
\end{table} 
\normalsize

\end{document}